\documentclass[runningheads]{llncs}
\usepackage{graphicx}
\usepackage{amsmath,amssymb} 
\usepackage{color}
\usepackage[width=122mm,left=12mm,paperwidth=146mm,height=193mm,top=12mm,paperheight=217mm]{geometry}

\usepackage{algorithm}
\usepackage{algpseudocode}
\usepackage{bbm}
\usepackage{bm}
\usepackage{subcaption}
\usepackage{url}

\newcommand\blfootnote[1]{%
  \begingroup
  \renewcommand\thefootnote{}\footnote{#1}%
  \addtocounter{footnote}{-1}%
  \endgroup
}

\DeclareMathOperator{\track}{Track}
\DeclareMathOperator*{\argmin}{argmin}
\DeclareMathOperator*{\argmax}{argmax}

\usepackage{hyperref}
\begin{document}
\pagestyle{headings}
\mainmatter

\title{Meta-Tracker: Fast and Robust Online Adaptation for Visual Object Trackers} 
\titlerunning{Meta-Tracker: Fast and Robust Online Adaptation}

\author{Eunbyung Park \quad \quad Alexander C. Berg}
\institute{Department of Computer Science,\\
    University of North Carolina at Chapel Hill\\
    \email{\{eunbyung,aberg\}@cs.unc.edu}}

\authorrunning{Eunbyung Park \& Alexander C. Berg}

\maketitle

\begin{abstract}
This paper improves state-of-the-art visual object trackers that use online adaptation. Our core contribution is an offline meta-learning-based method to adjust the initial deep networks used in online adaptation-based tracking. The meta learning is driven by the goal of deep networks that can quickly be adapted to robustly model a particular target in future frames. Ideally the resulting models focus on features that are useful for future frames, and avoid overfitting to background clutter, small parts of the target, or noise. By enforcing a small number of update iterations during meta-learning, the resulting networks train significantly faster. We demonstrate this approach on top of the high performance tracking approaches: tracking-by-detection based MDNet~\cite{nam-cvpr16-MDNet} and the correlation based CREST~\cite{song-iccv17-CREST}. Experimental results on standard benchmarks, OTB2015~\cite{otb} and VOT2016~\cite{vot2016}, show that our meta-learned versions of both trackers improve speed, accuracy, and robustness.

\end{abstract}

\section{Introduction}
\vspace{-2mm}
\blfootnote{The code, raw results and pretrained models are available at \url{https://github.com/silverbottlep/meta_trackers}}

Visual object tracking is a task that locates target objects precisely over a sequence of image frames given a target bounding box at the initial frame. In contrast to other object recognition tasks, such as object category classification and detection, in visual object tracking, instance-level discrimination is an important factor. For example, a target of interest could be one particular person in a crowd, or a specific product (e.g. coke can) in a broader category (e.g. soda cans). Therefore, an accurate object tracker should be capable of not only recognizing generic objects from background clutter and other categories of objects, but also discriminating a particular target among similar distractors that may be of the same category. Furthermore, the model learned during tracking should be flexible to account for appearance variations of the target due to viewpoint change, occlusion, and deformation.

One approach to these challenges is applying online adaptation. The model of the target during tracking, e.g. DCF (discriminative correlation filter) or binary classifier (the object vs backgrounds), is initialized at the first frame of a sequence, and then updated to be adapted to target appearance in subsequent frames~\cite{nam-cvpr16-MDNet,song-iccv17-CREST,DanelljanECCV2016,DanelljanCVPR2017,henriques2015tracking,Ma-ICCV-2015,kalal-tpami-2010,Bolme-cvpr-2010}. With the emergence of powerful generic deep-learning representations, recent top performing trackers now leverage the best of both worlds: deep learned features and online adaptation methods. Offline-only trackers trained with deep methods have also been suggested, with promising results and high speed, but with a decrease in accuracy compared to state-of-the-art online adaptive trackers~\cite{bertinetto2016fully,held2016learning,TaoCVPR2016}, perhaps due to difficulty finely discriminating specific instances in videos.

A common practice to combine deep learning features and online adaptation is to train a target model on top of deeply learned features, pre-trained over a large-scale dataset. These pre-trained features have proven to be a powerful and broad representation that can recognize many generic objects, enabling effective training of target models to focus on the specified target instance. Although this type of approach has shown the best results so far, there remains several important issues to be resolved. 

First, very few training examples are available. We are given a single bounding box for the target in the initial frame. In subsequent frames, trackers collect additional images, but many are redundant since they are essentially the same target and background. Furthermore, recent trends towards building deep models for target appearance~\cite{nam-cvpr16-MDNet,song-iccv17-CREST} make the problem more challenging since deep models are known to be vulnerable to overfitting on small datasets. As a consequence, a target model trained on top of deeply learned features sometimes suffers because it overfits to background clutter, small parts or features of the target, or noise. Many recent studies have proposed various methods to resolve these issues. Some include using a large number of positive and negative samples with aggressive regularizers~\cite{nam-cvpr16-MDNet}, factorized convolution~\cite{DanelljanCVPR2017}, spatio-residual modules~\cite{song-iccv17-CREST}, or incorporating contextual information~\cite{Mueller-cvpr-2017}. 

Second, most state-of-the-art trackers spend a significant amount of time on the initial training stage~\cite{nam-cvpr16-MDNet,song-iccv17-CREST,DanelljanCVPR2017}. Although many works have proposed fast training methods~\cite{DanelljanCVPR2017,henriques2015tracking}, this still remains a bottleneck. In many practical applications of object tracking, such as surveillance, real-time processing is required. Depending on the application, falling behind on the initial frame could mean failure on the whole task. On the other hand, an incompletely trained initial target model could affect performance on future frames, or in the worst case, result in failures on all subsequent frames. Therefore, it is highly desirable to obtain the robust target model very quickly at the initial frame.

In this work, we propose a generic and principled way of tackling these challenges. Inspired by recent meta-learning (learning to learn) studies~\cite{Finn-icml-2017,Ravi-iclr-2017,Andrychowicz-nips-2016,santoro-icml-16,Li-arxiv-2017,shedivat-iclr-2018}, we seek to learn how to obtain the target model. The key idea is to train the target model in a way that  generalizes well over future frames. In all previous works~\cite{nam-cvpr16-MDNet,song-iccv17-CREST,DanelljanECCV2016,DanelljanCVPR2017,henriques2015tracking,Ma-ICCV-2015,kalal-tpami-2010,Bolme-cvpr-2010}, the target model is trained to minimize a loss function on the current frame. Even if the model reaches an optimal solution, it does not necessarily mean it would work well for future frames. Instead, we suggest to use error signals from future frames. During the meta-training phase, we aim to find a generic initial representation and gradient directions that enable the target model to focus on features that are useful for future frames. Also, this meta-training phase helps to avoid overfitting to distractors in the current frame. In addition, by enforcing the number of update iterations during meta-training, the resulting networks train significantly faster during the initialization.

Our proposed approach can be applied to any learning based tracker with minor modifications. We select two {\em state-of-the-art trackers}, MDNet~\cite{nam-cvpr16-MDNet}, from the classifier based tracker (tracking-by-detection) category, and CREST~\cite{song-iccv17-CREST}, a correlation based tracker. Experimental results show that our meta-learned version of these trackers can adapt very quickly---just one iteration---for the first frame while improving accuracy and robustness. Note that this is done even without employing some of the hand engineered training techniques, sophisticated architectural design, and hyperparameter choices of the original trackers. In short, we present an easy way to make very good trackers even better without too much effort, and demonstrate its success on two different tracking architectures, indicating potentially general applicability.

\vspace{-3mm}
\section{Related Work}
\vspace{-3mm}

\noindent\textbf{Online trackers}: Many online trackers use correlation filters as the back-bone of the algorithms due to its computational efficiency and discriminative power. From the early success of the MOSSE tracker~\cite{Bolme-cvpr-2010}, a large number of variations have been suggested. \cite{henriques2015tracking} makes it more efficient by taking advantage of circulant matrices, further improved by resolving artificial boundary issues~\cite{danelljan-iccv-2015,Galoogahi-cvpr-2015}. Many hard cases have been tackled by using context information~\cite{Mueller-cvpr-2017,zhang-eccv-2014}, short and long-term memory~\cite{Ma-cvpr-2015,Hong-cvpr-2015}, and scale-estimation~\cite{danelljan-bmvc-2014}, just to name a few. Recently, deep learning features have begun to play an important role in correlation filters~\cite{nam-cvpr16-MDNet,song-iccv17-CREST,DanelljanECCV2016,DanelljanCVPR2017,Ma-ICCV-2015,Valmadre-cvpr-2017,Li-bmvc-2014}. On the other hand, tracking-by-detection approaches typically learn a classifier to pick up the positive image patches wrapping around the target object. Pioneered by~\cite{kalal-tpami-2010}, many learning techniques have been suggested, e.g. multiple instance learning~\cite{Babenko-tpami-2010}, structured output SVMs~\cite{Hare-tpami-2015}, online boosting~\cite{Grabner-eccv-2008}, and model ensembles~\cite{Bai-iccv-2013}. More recently, MDNet~\cite{nam-cvpr16-MDNet}, with deep features and a deep classifier, achieved significantly higher accuracy.

\noindent\textbf{Offline trackers}: Several recent studies have shown that we can build accurate trackers without online adaptation~\cite{bertinetto2016fully,held2016learning,TaoCVPR2016} due to powerful deep learning features. Siamese-style networks take a small target image patch and a large search image patch, and directly regress the target location~\cite{held2016learning} or generate a response map~\cite{bertinetto2016fully} via a correlation layer~\cite{Fischer-cvpr-2015}. In order to consider temporal information, recurrent networks have also been explored in~\cite{Kahou-arxiv-2015,Gan-arxiv-2015,Gordon-arxiv-2017,Yang-iccv-2017}. 

\noindent\textbf{Meta-learning}: This is an emerging field in machine learning and its applications. Although it is not a new concept~\cite{schumidhuber-1987,schumidhuber-1992,hochreiter-2001,thrun-1998}, many recent works have shown very promising results along with deep learning success. \cite{Andrychowicz-nips-2016,chen-icml-2017,Wichrowska2017a,Li-2017} attempted to replace hand-crafted optimization algorithms with meta-learned deep networks. \cite{Ravi-iclr-2017} took this idea into few shot or one shot learning problem. It aimed to learn optimal update strategies based on how accurate a learner can classify test images with few training examples when the learner follows the strategies from the meta-learner. Instead of removing existing optimization algorithms, \cite{Finn-icml-2017} focuses on learning initialization that are most suitable for existing algorithms. \cite{Li-arxiv-2017} further learns parameters of existing optimization algorithms along with the initialization. Unlike approaches introduced above, there also have been several studies to directly predict the model parameters without going through the optimization process~\cite{Yang-iccv-2017,Bertinetto-nips-2016,Wang-eccv-2016}. 

\vspace{-3mm}
\section{Meta-Learning for Visual Object Trackers} \label{sec:meta_training}
\vspace{-2mm}
In this section, we explain the proposed generalizable meta-training framework for visual object trackers. The details for applying this to each tracker are found in Section~\ref{sec:meta_trackers}.

\begin{figure*}[t]
\begin{center}
\includegraphics[width=\linewidth]{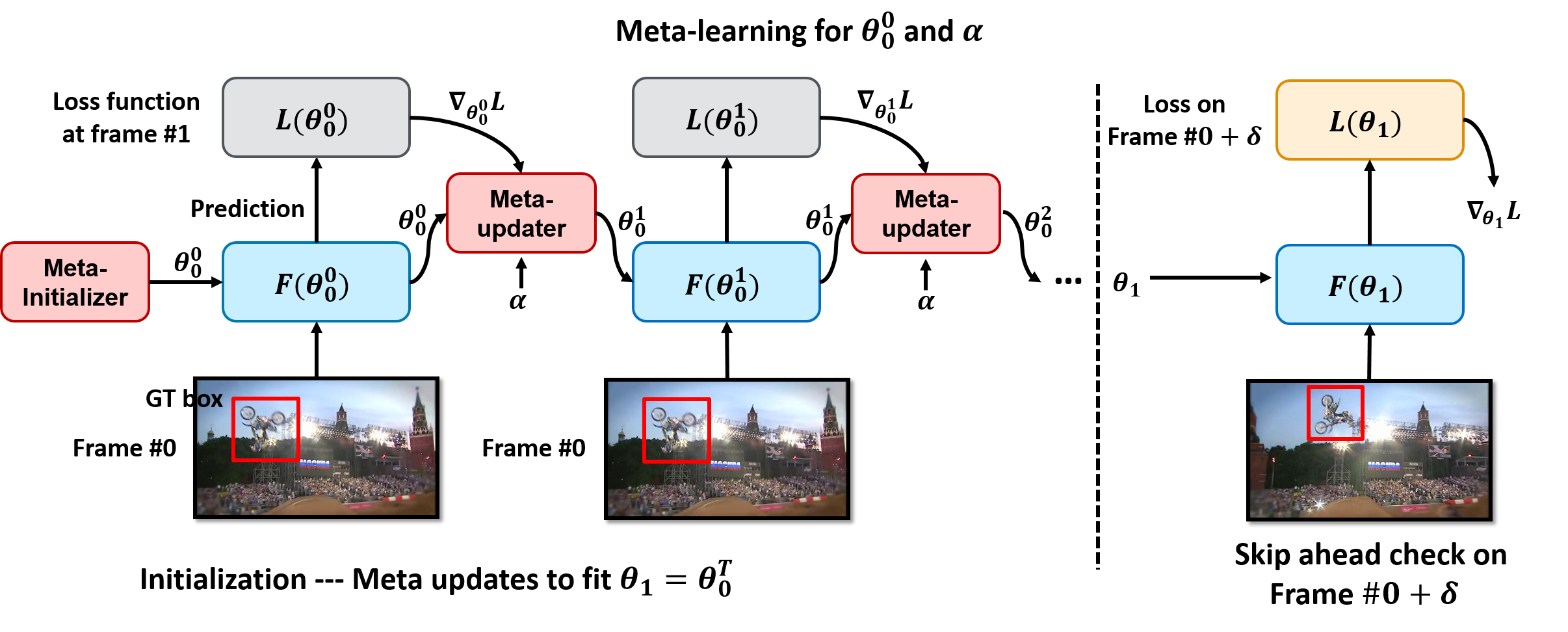}
\end{center}
\vspace{-5mm}
\caption{Our meta-training approach for visual object tracking: A computational graph for meta-training object trackers. For each iteration, it gets the gradient with respect to the loss after the first frame, and a meta-updater updates parameters of the tracker using those gradients. For added stability and robustness a final loss is computed using a future frame to compute the gradients w.r.t parameters of meta-initializer and meta-updater. More details in Section~\ref{sec:meta_training}.}
\label{fig:meta_training}
\vspace{-5mm}
\end{figure*}

\vspace{-3mm}
\subsection{Motivation}
\vspace{-2mm}
A typical tracking episode is as follows: The tracking model is adapted to a specified bounding box around the target in the initial frame of a sequence. Aggressive regularizers and fast optimization techniques are adopted to allow this adaptation/training to be done quickly so that the resulting model is  robust to target variations and environment changes. Then, the tracking model is used to predict the target location in subsequent frames. Predicted target locations and images are then stored in the database, and the models are regularly updated with collected data according to their own strategies. 

A key motivation is to incorporate these actual tracking scenarios into the meta-learning process. The eventual goal of trackers is to  predict the target locations in future frames. Thus, it would be desirable to learn trackers with this eventual goal. For example, if we could look at variations in future frames, then we could build more robust target models and prevent them from overfitting to the current target appearance or background clutter. We can take a step back and observe trackers running on videos, see if the trackers generalize well, and find a reason why they become distracted and adjust the adaptation procedure accordingly.

\vspace{-3mm}
\subsection{A general online tracker}
\vspace{-2mm}
This formulation of online tracking is made general in order to apply to a variety of trackers.  Consider the key operation in a tracker, $\hat{y}=F(x,\theta)$, that takes an input $x$, e.g. image patches around the target or a cropped image centered on putative target from an image $I$, and the tracker parameters $\theta$ and produces an estimate $\hat{y}$ of the label, e.g. a response map or a location in the frame that indicates the target position. For {\em initialization}, $x_0$ from the initial frame $I_0$ with specified $y_0$, we (approximately) solve for $\theta_1(x_0,y_0)$, or $\theta_1$ for brevity, with respect to a loss, $L\left({F\left({x_0,\theta_1}\right),y_0}\right)$ measuring how well the model predicts the specified label. For {\em updates} during tracking, we take the parameters $\theta_j$ from frame $j-1$ and find $\hat{y}_{j}=F(x_j,\theta_j)$, then find $\theta_{j+1}$ with respect to a loss. Then, we may incorporate transforming $\hat{y}_j$ into a specific estimate of the target location as well as temporal smoothing, etc. 
We can write the tracking process initialized with $x_0$ and $y_0$ in an initial frame and then proceeding to track and update for frames $I_1 \ldots I_n$ as $\track \left({\theta_1(x_0,y_0),I_1,\ldots,I_n}\right)$ and its output as $\hat{y}_{n}$ an estimate of the label in the $n$th frame (indicating target position) and $\theta_{n+1}$, the model parameters after the $n$th frame.



\vspace{-3mm}
\subsection{Meta-training algorithm}
\vspace{-2mm}

\begin{algorithm}[t]
\caption{Meta-training object trackers algorithm}
\label{alg:algorithm}
\textbf{Input} : Randomly initialized $\theta_0$ and $\alpha$, training dataset $D$\\
\textbf{Output} : $\theta_0^*$ and $\alpha^*$
\begin{algorithmic}[1]
\While{not converged}
\State $\textrm{grad}_{\theta_0}, \textrm{grad}_{\alpha} = \vec{0}$ \Comment{Initialize to zero vector}
\For{\textbf{all}  $k \in \{0,\dots, N_{\textrm{mini}}-1\}$}
\State $S, j, \delta \sim p(D)$ \Comment{Sample a training example}
\State $\theta^0_0 = \theta_0$
    \For{\textbf{all} $t \in \{0,\dots, T-1\}$}
    \State $\hat{y}_j = F(x_j,\theta^t_0)$ 
    \State $\theta^{t+1}_0 = \theta^{t}_0 - \alpha \odot \nabla_{\theta^t_0} L(y_j,\hat{y}_j;\theta^t_0)$ 
    \EndFor
\State $\theta_1 = \theta_0^T$
\State $\hat{y}_{j+\delta} = F(x_{j+\delta},\theta_1)$ \Comment{Apply to a future frame}
\State $\textrm{grad}_{\theta_0} = \textrm{grad}_{\theta_0} + \nabla_{\theta_0} L(y_{j+\delta},\hat{y}_{j+\delta})$ \Comment{Accumulate the gradients}
\State $\textrm{grad}_{\alpha} = \textrm{grad}_{\alpha} + \nabla_{\alpha} L(y_{j+\delta},\hat{y}_{j+\delta})$
\EndFor
\State $\theta_0 = \textrm{Optimizer}(\theta_0, \textrm{grad}_{\theta_0})$ \Comment{Update $\theta_0$}
\State $\alpha = \textrm{Optimizer}(\alpha, \textrm{grad}_{\alpha})$ \Comment{Update $\alpha$}
\EndWhile
\end{algorithmic}
\end{algorithm}

Our meta-training approach has two goals.  One is that initialization for a tracker on a sequence can be performed by starting with $\theta_0$ and applying one or a very small number of iterations of a update function $M$ parameterized by $\alpha$. Another goal is that the resulting tracker be accurate and robust on later frames.

The gradient-descent style update function $M$ is parameterized by $\alpha$:
\begin{equation} \label{eq:1}
M(\theta,\nabla_{\theta}L;\alpha) = \theta - \alpha \odot \nabla_{\theta}L\,,
\end{equation}
\noindent where $\alpha$ is the same size as the tracker parameters $\theta$~\cite{Li-arxiv-2017}, $L$ is a loss function, and $\odot$ is element-wise product. $\alpha$ could be a scalar value, which might be either learnable~\cite{shedivat-iclr-2018} or manually fixed~\cite{Finn-icml-2017}. We empirically found that having per parameter coefficients was the most effective in our settings.

Our meta-training algorithm is to find a good $\theta_0$ and $\alpha$ by repeatedly sampling a video, performing initialization, applying the learned initial model to a frame slightly ahead in the sequence, and then back-propagating to update $\theta_0$ and $\alpha$. Applying the initial model to a frame slightly ahead in the sequence has two goals, the model should be robust enough to handle more than frame-to-frame variation, and if so, this should make updates during tracking fast as well if not much needs to be fixed.

After sampling a random starting frame from a random video, we perform optimization for initialization starting with $\theta^0_0 = \theta_0$ given the transformed input and output pair, $(x_j,y_j)$. A step of optimization proceeds as
\begin{equation} \label{eq:2}
\theta^{i+1}_0 = M(\theta^{i}_0,\nabla_{\theta^{i}_0}L(y_j,F(x_j,\theta^{i}_0)))\,.
\end{equation}
\noindent This step can be repeated up to a predefined number of times $T$ to find, $\theta_1(x_j,y_j)=\theta^T_0$. Then, we randomly sample a future frame $I_{j+\delta}$ and evaluate the model trained on the initial frame on that future frame to produce: $\hat{y}_{j+\delta}=F(x_{j+\delta},\theta_1 )$.

The larger $\delta$, the larger target object variations and environment changes are incorporated into training process. Now, we can compute the loss based on the future frame and trained tracker parameters. The objective function is defined as
\begin{equation} \label{eq:3}
\theta_0^*, \alpha^* = \argmin_{\theta_0,\alpha} \mathbb{E}_{S, j, \delta} [ L(y_{j+\delta},\hat{y}_{j+\delta}) ]\,.
\end{equation}
We used the ADAM~\cite{adam} gradient descent algorithm to optimize. Note that $\theta_0$ and $\alpha$ are fixed across different episodes in a mini-batch, but $\theta_0^1,\dots,\theta_0^T$ are changed over every episode. To compute gradients of the objective function w.r.t $\theta_0$ and $\alpha$, it is required to compute higher-order gradients (the gradients of function of gradients). This type of computation has been exploited in recent studies~\cite{Finn-icml-2017,Maclaurin-icml-2015,metz-iclr-2017}. We can easily compute this thanks to automatic differentiation software libraries~\cite{pytorch}. More details are explained in Algorithm~\ref{alg:algorithm}.

\noindent\textbf{Update rules for subsequent frames.} Most online trackers, including the two trackers we meta-train (Section \ref{sec:meta_trackers}), update the target model regularly to adjust to new examples collected by itself during tracking. We could simply use meta-trained $\alpha$ to update the model, $\theta_j = \theta_{j-1} - \alpha \odot \nabla_{\theta_{j-1}}L$ (only one iteration presented for brevity). However, it often diverges on longer sequences or the sequences that have very small frame-to-frame variations. We believe this is mainly because we train $\alpha$ for fast adaptation at the initial frame, so the values of $\alpha$ are relatively large, which causes unstable convergence behavior (A similar phenomenon was reported in~\cite{shedivat-iclr-2018} albeit in a different context). Since $\alpha$ is stable when it teams up with $\theta_0$, we could define the update rules for subsequent frames as $\theta_j = \theta_0 - \alpha \odot \nabla_{\theta_0}L$, as suggested in~\cite{shedivat-iclr-2018}. We could also combine two strategies, $\theta_j = \beta(\theta_{j-1} - \alpha \odot \nabla_{\theta_{j-1}}L) + (1-\beta)(\theta_0 - \alpha \odot \nabla_{\theta_0}L)$. Although we could resolve unstable convergence behavior with these strategies, none of these performed better than simply searching for a single learning rate. Therefore, we find a learning rate for subsequent frames and then use existing optimization algorithms to update the models as was done in the original versions of the trackers.

\vspace{-3mm}
\section{Meta-Trackers} \label{sec:meta_trackers}
\vspace{-2mm}
In this section, we show how our proposed meta-learning technique can be realized in state-of-the-art trackers. We selected two different types of trackers, one from correlation based trackers, CREST~\cite{song-iccv17-CREST}, and one from tracking-by-detection based trackers MDNet~\cite{nam-cvpr16-MDNet}.

\vspace{-3mm}
\subsection{Meta-training of correlation based tracker} \label{meta_crest}
\vspace{-2mm}

\begin{figure}[t]
\begin{minipage}[b]{0.48\textwidth}
\includegraphics[width=\linewidth]{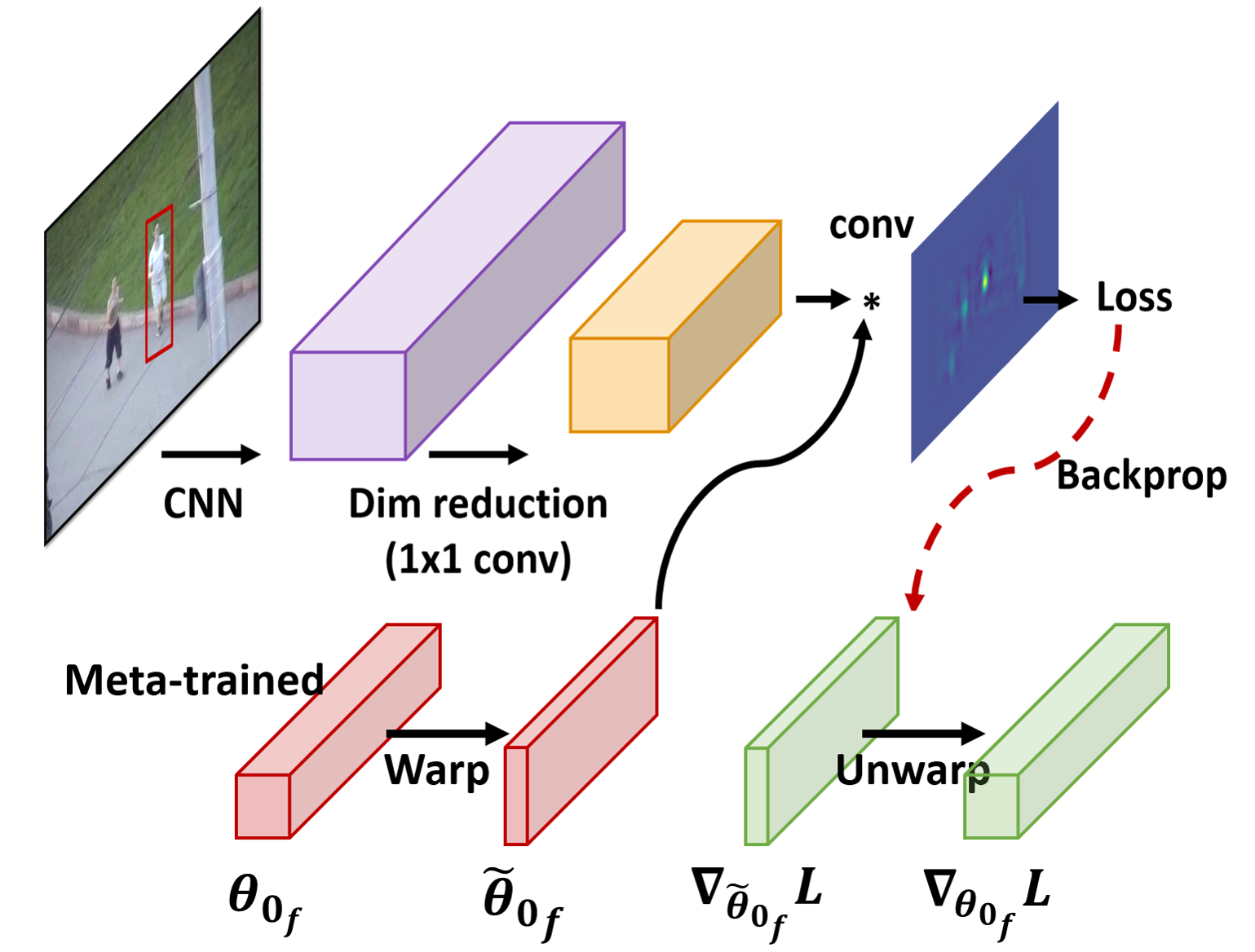}
\subcaption{MetaCREST}\label{fig:meta_crest}
\end{minipage}
\begin{minipage}[b]{0.48\textwidth}
\includegraphics[width=\linewidth]{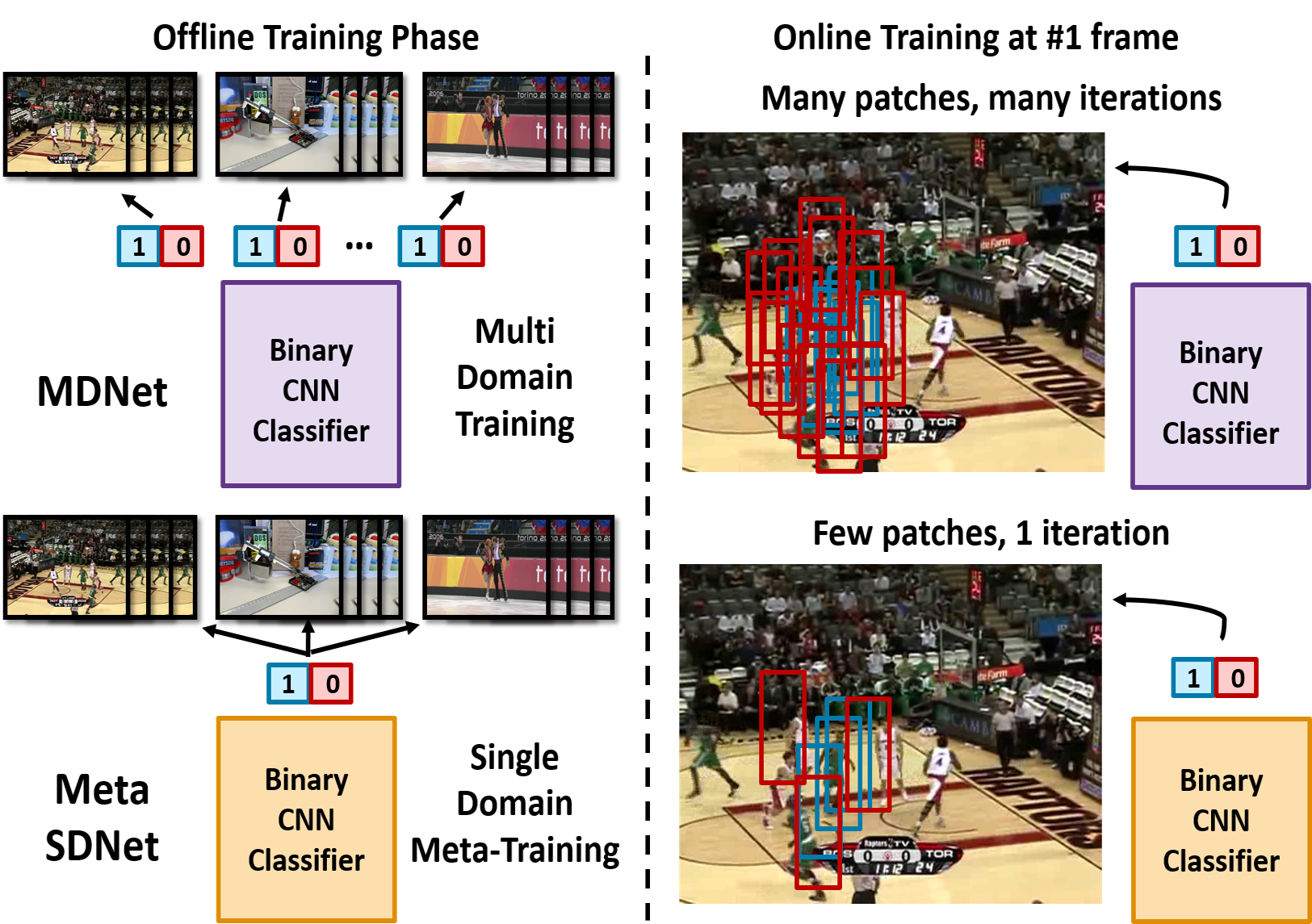}
\subcaption{MDNet vs MetaSDNet}\label{fig:meta_sdnet}
\end{minipage}
\vspace{-5mm}
\end{figure}

\noindent\textbf{CREST.} A typical correlation filter objective is defined as follows.
\begin{equation} \label{eq:corr_loss}
\argmin_{f} || y - \Phi(x)*f ||^2 + \lambda||f||^2\,,
\end{equation}
where $f$ is the correlation filter, $*$ is the convolution operation, and $\Phi$ is a feature extractor, e.g. CNN. $x$ is a cropped image centered on the target, and $y \in \mathbb{R}^{H \times W}$, is a gaussian shaped response map, where $H$ and $W$ are height and width, respectively. The cropped image is usually larger than the target object so that it can provide enough background information. Once we have the correlation filter trained, target localization at a new future frame is simply finding the coordinates $(h,w)$ that has the maximum response value. 

\begin{equation}
\argmax_{(h,w)} \hat{y}(h,w)\,,
\end{equation}
where $\hat{y} = \Phi(x_\textrm{new})*f$, and $\hat{y}(h,w)$ represents the element of $\hat{y}$ in $(h,w)$ coordinates. CREST used a variation of the correlation filter objective, defined as
\begin{equation} \label{eq:crest_loss}
\sum_{(h,w) \in P} \frac{1}{|P|} ( e^{y(h,w)} \lvert y(h,w) - \hat{y}(h,w) \rvert )^2 + \lambda||f||^2\,,
\end{equation}
where $P = \{(h,w) \mid \lvert y(h,w)-\hat{y}(h,w) \rvert > 0.1 \}$. This would encourage the model to focus on parts that are far from the ground truth values.

By reformulating the correlation filter as a convolutional layer, it can be effectively integrated into an CNN framework~\cite{song-iccv17-CREST}. This allows us to add new modules easily, since the optimization can be nicely done with standard gradient descent in end-to-end fashion. They inserted spatio-temporal residual modules to avoid target model degradation by large appearance changes. They also devised sophisticated initialization, learning rates, and weight decay regularizers, e.g. 1000 times larger weight decay parameter on spatio-temporal residual modules. Without those bells and whistles, we aim to learn a robust single layer correlation filter via proposed meta-learning process. There are two important issues for plugging CREST tracker into proposed meta-training framework, and we present our solutions in following sections.

\noindent\textbf{Meta-learning dimensionality reduction.} CREST used PCA to reduce the number of channels of extracted CNN features, from 512 to 64. This not only reduces computational cost, but also it helps to increase robustness of the correlation filter. PCA is performed at the initial frame and learned projection matrix are used for the rest of the sequence. This becomes an issue when meta-training the correlation filter. We seek to find a global initialization of the correlation filter for the all targets from different episodes. However, PCA would change the basis for every sequences, which makes impossible to obtain a global initialization in projected feature spaces that are changing every time. We propose to learn to reduce dimensions of the features. In CREST, we can insert 1x1 convolution layer right after the feature extraction, the weights of this layer are also meta-learnable and jointly trained during the meta-learning process along with the correlation filter. $\theta_0$ in proposed meta-training framework, therefore, consists of $\theta_{{0}_d}$ and $\theta_{{0}_f}$, the parameters of dimensionality reduction and the correlation filter, respectively.

\noindent\textbf{Canonical size initialization.} The size of the correlation filter varies depending on the target shape and size. In order to meta-train a fixed size initialization of the correlation filter $\theta_{{0}_f}$, we should resize all objects to the same size and same aspect ratio. However, it introduces distortion of the target and has been known to degrade recognition performance~\cite{liu-eccv-2016,ren-nips-2016}. In order to fully make use of the power of the correlation filter, we propose to use canonical size initialization and its size and aspect ratio are calculated as a mean of the objects in the training dataset. Based on canonical size initialization, we warp it to the specific size taylored to the target object for each tracking episodes, $\tilde{\theta}_{{0}_f}=\textrm{Warp}(\theta_{{0}_f})$. We used differentiable bilinear sampling method~\cite{jaderberg-nips-2015} to pass through gradients all the way down to $\theta_{{0}_f}$.

Putting it all together, $F(x_j,\theta)$ in our proposed meta-training framework for CREST, now takes an input a cropped image $x_j$ from an input frame $I_j$, pass it through a CNN feature extractor followed by dimensionality reduction (1x1 convolution with the weight $\theta_{{0}_d}$). Then, it warps the correlation filter $\theta_{{0}_f}$, and finally apply warped correlation filter $\tilde{\theta}_{{0}_f}$ to produce a response map $\hat{y}_j$ (Figure~\ref{fig:meta_crest}).

\vspace{-3mm}
\subsection{Meta-training of tracking-by-detection tracker} \label{meta_mdnet}
\vspace{-2mm}

\noindent\textbf{MDNet.} MDNet is based on a binary CNN classifier consisting of a few of convolutional layers and fully connected layers. In the offline phase, it uses a multi-domain training technique to pre-train the classifier. At the initial frame, it randomly initializes the last fully connected layer, and trains around 30 iterations with a large number of positive and negative patches (Figure~\ref{fig:meta_sdnet}). Target locations in the subsequent frames are determined by average of bounding box regression outputs of top scoring positive patches. It collects positive and negative samples during the tracking process, and regularly updates the classifier. Multi-domain pre-training was a key factor to achieve robustness, and they used an aggressive dropout regularizer and different learning rates at different layers to further avoid overfitting to current target appearance. Without those techniques (the multi-domain training and regularizers), we aim to obtain robust and quickly adaptive classifier solely resting on the proposed meta-learning process.

\noindent\textbf{Meta-training.} It can be also easily pluged into the proposed meta-leraning framework. $F(x_j;\theta)$ takes as input image patches $x_j \in \mathbb{R}^{N \times D}$ from an input frame $I_j$ (and $y_j \in \{0,1\}^N$ is the corresponding labels), where $D$ is the size of the patches and $N$ is the number of patches. Then, the patches go through a CNN classifier, and the loss function $L$ is a simple cross entropy loss $-\sum_{k=1}^N y_j^k\log(F^k(x_j; \theta))$.

\noindent\textbf{Label shuffling.} Although a large-scale video detection dataset contains rich variation of objects in videos, the number of objects and categories are limited compared to other still image datasets. This might lead a deep CNN classifier to memorize all object instances in the dataset and classify newly seen objects as backgrounds. In order to avoid this issue, we adopted the label shuffling trick, suggested in~\cite{santoro-icml-16}. Every time we run a tracking episode, we shuffle the labels, meaning sometimes labels of positive patches become 0 instead of 1, negative patches become 1 instead of 0. This trick encourages the classifier to learn how to distinguish the target objects from background by looking at current training examples, rather than memorizing specific targets appearance. 

\vspace{-3mm}
\section{Experiments}
\vspace{-2mm}

\subsection{Experimental setup}

\noindent\textbf{VOT2016.} It contains 60 videos (same videos from VOT 2015~\cite{vot2015}). Trackers are automatically reinitialized once it drifts off the target: zero overlap between predicted bounding box and the ground truth. In this reset-based experiments, three primary measures have been used, (i) \textit{accuracy}, (ii) \textit{robustness} and (iii) \textit{expected average overlap (EAO)}. The accuracy is defined as average overlap during successful tracking periods. The robustness is defined as how many times the trackers fail during tracking. The expected average overlap is an estimator of the average overlap a tracker is expected to attain on a large collection of short-term sequences.

\noindent\textbf{OTB2015.} It consists of 100 fully annotated video sequences. Unlike VOT2016, the one pass evaluation (OPE) is commonly used in OTB dataset (no restart at failures). The precision plots (based on the center location error) and the success plots (based on the bounding box overlap) are used to access the tracker performance.

\noindent\textbf{Dataset for meta-training.} We used a large scale video detection dataset~\cite{ILSVRC15} for meta-training both trackers. It consists of 30 object categories, which is a subset of 200 categories in the object detection dataset. Since characteristics of the dataset are slightly different from the object tracking dataset, we sub-sampled the dataset. First, we picked a video frame that contains a target object whose size is not larger than 60\% of the image size. Then, a training video sequence is constructed by sampling all subsequent frames from that frame until the size of the target object reaches 60\%. We ended up having 718 video sequences. In addition, for the experiments on OTB2015 dataset, we also used an additional 58 sequences from object tracking datasets in VOT2013,VOT2014,and VOT2015~\cite{vot2016}, excluding the videos included in OTB2015, following MDNet's approach~\cite{nam-cvpr16-MDNet}. These sequences were selected in the mini-batch selection stage with the probability 0.3. Similarly, we used 80 sequences from OTB2015, excluding the videos in VOT2016 for the experiments on VOT2016 dataset.

\noindent\textbf{Baseline implementations.} We selected two trackers, MDNet~\cite{nam-cvpr16-MDNet} and CREST~\cite{song-iccv17-CREST}. For CREST, we re-implemented our own version in python based on publicly released code written in MATLAB. We meta-trained our version. For MDNet, the authors of MDNet provide two different source codes, written in MATLAB and python, respectively. We used the latter one and called it as pyMDNet or pySDNet, depending on pre-training methods. We meta-trained pySDNet, and call it as MetaSDNet. Note that overall accuracy of pyMDNet is lower than MDNet on OTB2015 (.652 vs .678 in success rates with overlap metric). For fair comparison, we compared our MetaSDNet to pyMDNet.

\noindent\textbf{Meta-training details.} In MetaSDNet, we used the first three conv layers from pre-trained vgg16 as feature extractors. During meta-training, we randomly initialized the last three fc layers, and used Adam as the optimizer with learning rate 1e-4. We only updated the last three fc layers for the first 5,000 iterations and trained all layers for the rest of iterations. The learning rate was reduced to 1e-5 after 10,000 iterations, and we trained up to 15,000 iterations. For $\alpha$, we initialized to 1e-4, and also used Adam with learning rate 1e-5, then was decayed to 1e-6 after 10,000 iterations. We used mini-batch size $N_{\textrm{mini}}=8$. For the meta-update iteration $T$, larger $T$ gave us only small improvement, so we set to 1. In MetaCREST, we randomly initialized $\theta_0$ and also used Adam with learning rate 1e-6. For $\alpha$, we initialized to 1e-6, and learning rate of Adam was also set to 1e-6. $N_{\textrm{mini}}=8$ and meta-training iterations was 10,000 (at 50,000 iterations, the learning rate was reduced to 1e-7). We used same hyper-parameters for both OTB2015 and VOT2016 experiments. For other hyper-parameters, we mostly followed the ones in the original trackers. For more details, the code and raw results will be released.

\vspace{-3mm}
\subsection{Experimental results}
\vspace{-2mm}
\subsubsection{Quantitative evaluation.} 

\begin{table}[t]
\caption{Quantitative results on VOT2016 dataset. The numbers in legends represent the number of iterations at the initial frame. EAO (expected average overlap) - 0 to 1 scale, higher is better. A (Accuracy) - 0 to 1 scale, higher is better. R (Robustness) - 0 to N, lower is better. We ran each tracker 15 times and reported averaged scores following VOT2016 convention.}
\label{table:vot2016_results}
\begin{minipage}[b]{0.5\textwidth}
\centering
\begin{tabular}{l|c|c|c}
              & EAO & Acc & R \\ \hline
MetaCREST-01 & \textbf{0.317}  & \textbf{0.519}  & \textbf{0.932} \\
CREST       & 0.283  & 0.514  & 1.083 \\
CREST-Base  & 0.249  & 0.502  & 1.383 \\
CREST-10    & 0.252  & 0.509  & 1.380 \\
CREST-05    & 0.262  & 0.510  & 1.298 \\
CREST-03    & 0.262  & 0.514  & 1.283 \\
CREST-01    & 0.259  & 0.505  & 1.277 \\
\end{tabular}
\subcaption{The results of MetaCREST}
\label{table:vot2016_crest}
\end{minipage}
\begin{minipage}[b]{0.5\textwidth}
\centering
\begin{tabular}{l|c|c|c}
              & EAO & Acc & R \\ \hline
MetaSDNet-01 & \textbf{0.314} & 0.526 & \textbf{0.934} \\
pyMDNet-30   & 0.304 & \textbf{0.540} & 0.943 \\
pyMDNet-15   & 0.299 & 0.541 & 0.977 \\
pyMDNet-10   & 0.291 & 0.535 & 0.989 \\
pyMDNet-05   & 0.254 & 0.523 & 1.198 \\
pyMDNet-03   & 0.184 & 0.488 & 1.703 \\
pyMDNet-01   & 0.119 & 0.431 & 2.733 \\
\end{tabular}
\subcaption{The results of MetaSDNet}
\label{table:vot2016_mdnet}
\end{minipage}
\vspace{-5mm}
\end{table}

\begin{figure*}[t]
\begin{center}
\includegraphics[width=0.7\linewidth]{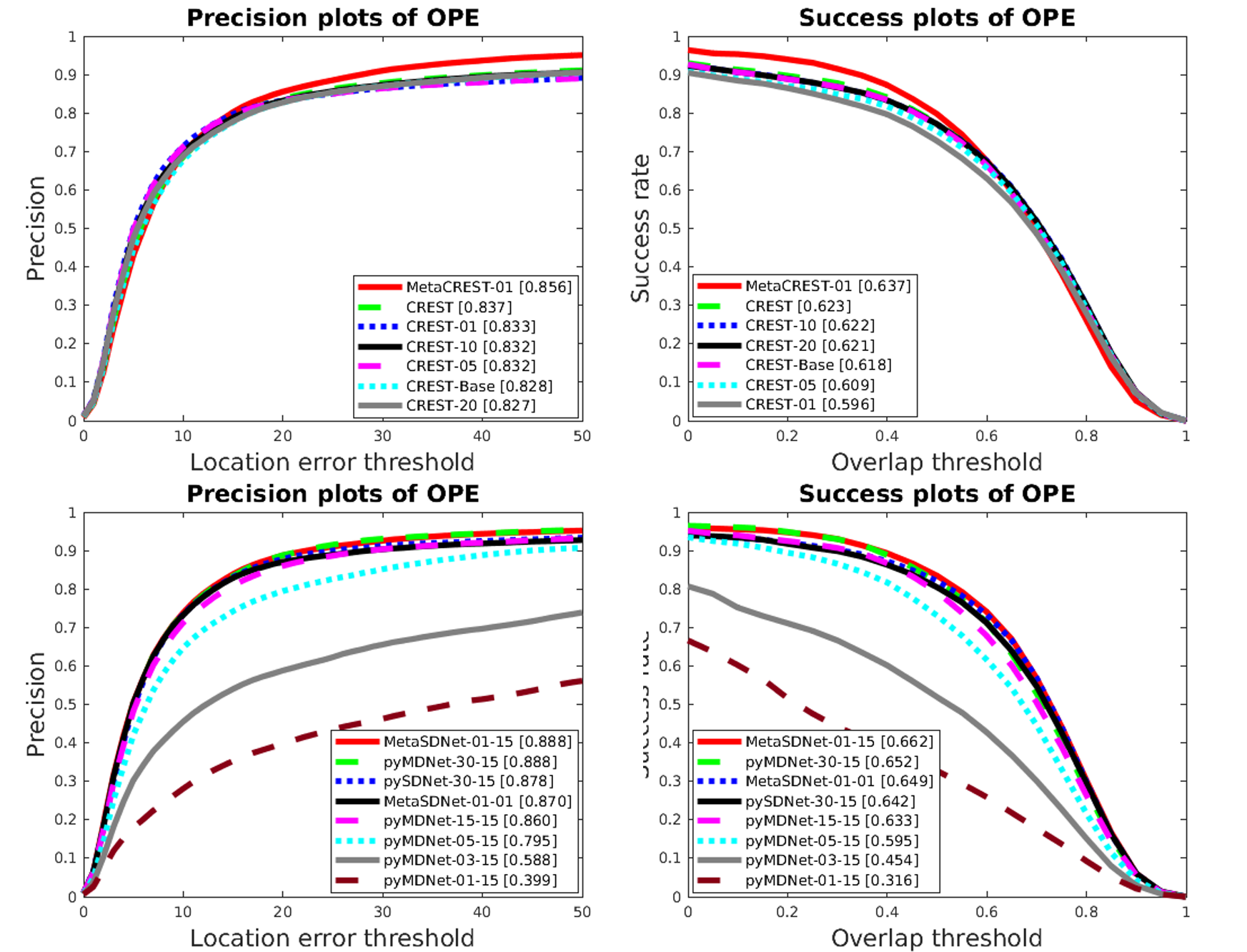}
\end{center}
\vspace{-5mm}
\caption{Precision and success plots over 100 sequences in OTB2015 dataset with one-pass evaluation (OPE). For CREST (top row),The numbers in legends represent the number of iterations at the initial frame, and all used 2 iterations for the subsequent model updates. For MDNet experiments (bottom row), 01-15 means, 1 training iterations at the initial frame and 15 training iterations for the subsequent model updates.}
\label{fig:otb100_results}
\vspace{-5mm}
\end{figure*}

Table~\ref{table:vot2016_results} shows quantitative results on VOT2016. In VOT2016, EAO is considered as the main metric since it consider both accuracy and robustness. Our meta-trackers, both MetaCREST and MetaSDNet, consistently improved upon their counterparts by significant margins. Note that this is the improvement without their advanced techniques, e.g. pyMDNet with specialized multi-domain training and CREST with spatio-temporal residual modules. The performances of the accuracy metric are not very different than the original trackers. Because it computes the average overlap by only taking successful tracking periods into account, we did not change other factors that might affect the accuracy in the original trackers, e.g. scale estimation. Quantitative results on OTB2015 are depicted in Figure~\ref{fig:otb100_results}. Both of MetaSDNet and MetaCREST also improved upon their counterparts in both precision and success plots with only one iteration for initialization. Detailed results of individual sequences in both of VOT2016 and OTB2015 are presented in Appendix.

We require only one iteration at the initial frame to outperform the original trackers. We also performed the experiments with more than one iteration, but the performance gain was not significant. On the other hand, MDNet takes 30 iterations to converge at the initial frame as reported in their paper, and fewer iterations caused serious performance degradation. This confirms that getting a robust target model at the initial frame is very important for subsequent frames. For CREST, performance drop was not significant as MDNet, but it was still more than 10 iterations to reach to its maximum performance. MDNet updates the model 15 iterations for subsequent frames at every 10 frames regularly (or when it failed, meaning its score is below a predefined threshold).

\begin{table}[t]
\caption{Speed and performance of the initialization: The right table shows the losses of estimated response map in MetaCREST. The left table shows the accuracy of image patches in MetaSDNet. B (Before) - the performance of the initial frame before training, A (After) - the performance of the initial frame after training, LH (Lookahead) - the performance of next 5 frames after training, Time - wall clock time to train in seconds}
\label{table:speed_accuracy}
\begin{minipage}[b]{0.5\textwidth}
\centering
\begin{tabular}{l|ccc|c}
              & B & A & LH & Time(s) \\ \hline
MetaCREST-01  & 0.48 & 0.04 & \textbf{0.05} & \textbf{0.090} \\
CREST-01    & 0.95  & 0.82  & 0.87 & 0.395 \\
CREST-03    & 0.95  & 0.62  & 0.75 & 0.424 \\
CREST-05    & 0.95  & 0.45  & 0.63 & 0.550 \\
CREST-10    & 0.95  & 0.24  & 0.40 & 0.668 \\
CREST-20    & 0.95  & 0.18  & 0.31 & 1.048 \\
CREST-65    & 0.95  & 0.01  & 0.30 & 1.529 \\
\end{tabular}
\end{minipage}
\begin{minipage}[b]{0.5\textwidth}
\centering
\begin{tabular}{l|ccc|c}
              & B & A & LH & Time(s) \\ \hline
MetaSDNet-01  & 0.50  & 0.98  & \textbf{0.97} & \textbf{0.124} \\
pyMDNet-01    & 0.51  & 0.56  & 0.56 & 0.123 \\
pyMDNet-03    & 0.51  & 0.79  & 0.78 & 0.373 \\
pyMDNet-05    & 0.51  & 0.84  & 0.84 & 0.656 \\
pyMDNet-10    & 0.51  & 0.95  & 0.93 & 1.171 \\
pyMDNet-15    & 0.51  & 0.97  & \textbf{0.97} & 1.819 \\
pyMDNet-30    & 0.51  & 0.99  & \textbf{0.98} & 3.508 \\
\end{tabular}
\end{minipage}
\end{table}

\vspace{-3mm}
\subsubsection{Speed and performance of the initialization.}
We reported the wall clock time speed at the initial frame in Table \ref{table:speed_accuracy}, on a single TITAN-X GPU. In CREST, in addition to feature extraction, there are two more computational bottlenecks. The first is the convolution with correlation filters. Larger objects means larger filters and more computations. We reported average time across all 100 sequences. Another heavy computation comes from PCA at the initial frame. It also depends on the size of the objects. Larger objects lead to larger center cropped images, features, and more computation in PCA.

MDNet requires many positive and negative patches, and also many model update iterations to converge. A large part of the computation comes from extracting CNN features for every patch. MetaSDNet needs only a few training patches and can achieve 30x speedup (0.124 vs 3.508), while improving accuracy. If we used more compact CNNs for feature extractions, the speed could have been in the range of real-time processing. For subsequent frames in MDNet, model update time is of less concern because MDNet only updates the last 3 fully connected layers, which are relatively faster than feature extractors. The features are extracted at every frame, stored in a database, and update the model every 10 frames. Therefore, the actual computation is well distributed across every frames.

We also showed the performance of the initialization to see the effectiveness of our approach (in Table \ref{table:speed_accuracy}. We measured the performance with learned initialization. After initial training, we measure the performance on the first frame and 5 future frames to see generalizability of trackers. MetaSDNet achieved very high accuracy after only one iteration, but accuracy of pyMDNet after one iteration was barely above guessing (guessing is 50\% and all negative prediction is 75\% accuracy since sampling ratio was 1:3 between positive and negative samples). The effectiveness is more apparent in MetaCREST. MetaCREST-01 without any updates gave already close performance to CREST-05 after training (0.48 vs 0.45). In original CREST tracker, they train the model until it reaches a loss of 0.02, which corresponds to an average 65 iterations. However, its generalizability at future frames is limited compared to ours (.05 vs .30). Although this is not directly proportional to eventual tracking performance, we believe this is clear evidence that our meta-training algorithm based on future frames is indeed effective, as also supported by overall tracking performance.

\subsubsection{Visualization of response maps.}

\begin{figure*}[t]
\begin{center}
\includegraphics[width=\linewidth]{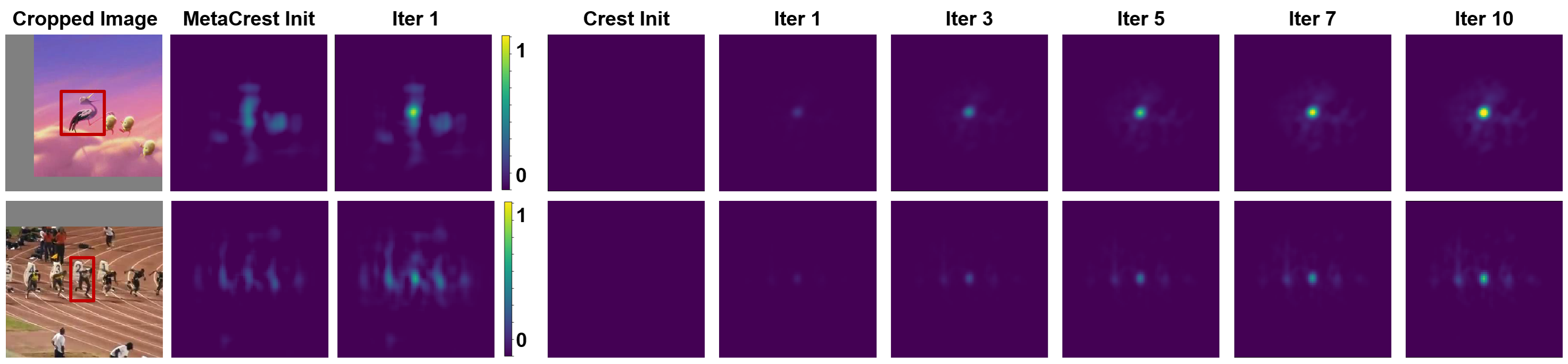}
\end{center}
\vspace{-5mm}
\caption{Visualizations of response maps in CREST: Left three columns represents the image patch at the initial frame, response map with meta-learned initial correlation filters $\theta_{{0}_f}$, response map after updating 1 iteration with learned $\alpha$, respectively. The rest of seven columns on the right shows response maps after updating the model up to 10 iterations.}
\label{fig:meta_crest_response}
\vspace{-5mm}
\end{figure*}

We visualized response maps in MetaCREST at the initial frame (Figure~\ref{fig:meta_crest_response}). A meta-learned initialization, $\theta_0$ should be capable of learning generic objectness or visual saliency. At the same time, it should not be instance specific. It turns out that is the case. The second column in Figure~\ref{fig:meta_crest_response} shows response maps by applying correlation filters to the cropped image (first column) with $\theta_0$. Without any training, it already generates high response values on some locations where there are objects. But, more importantly, there is no clear maximum. After one iteration, the maximum is clearly located at the center of the response map. In contrast to MetaCREST, CREST consumes more iterations to produce high response values on the target.

\begin{figure*}[t]
\begin{center}
\includegraphics[width=\linewidth]{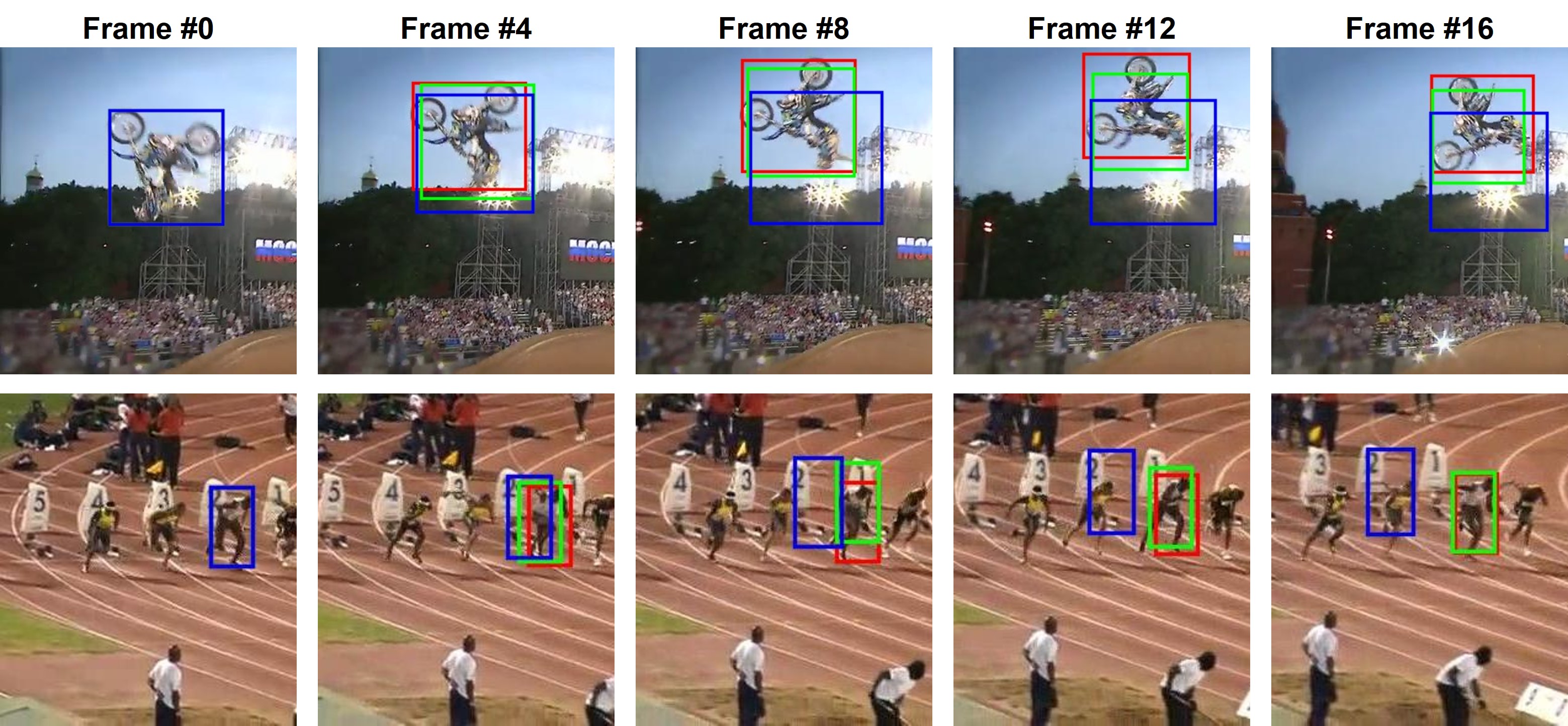}
\end{center}
\caption{Qualitative examples: tracking results at early stage of MotorRolling (top) and Bolt2 (bottom) sequences in OTB2015 dataset. Color coded boxes: ground Truth (Red), MetaCREST-01 (Green) and CREST (Blue).}
\label{fig:qualitative_examples}
\vspace{-5mm}
\end{figure*}

\subsubsection{Qualitative examples of robust initialization.} In Figure~\ref{fig:qualitative_examples}, 
we present some examples where MetaCrest overcomes some of the issues in the original CREST. 
In MotorRolling sequence (top row), CREST was distracted by a horizontal line from the forest in the background. CREST easily reached to 0.0000 loss defined in Equation~\ref{eq:crest_loss} at the initial frame, as opposed to 0.1255 in MetaCREST. This is a strong evidence that an optimal solution does not necessarily mean good generalizability on future frames. In contrast, MetaCREST, generalizes well to future frames, despite not finding an optimal solution at the current frame. In Bolt2 sequence (bottom row), CREST also reached to 0.0000 loss (vs 0.0534 in MetaCREST). In a similar way, a top left part in the bounding box was the distractor. MetaCREST could easily ignore the background clutter and focused on the object in the center of the bounding box.

\vspace{-3mm}
\section{Conclusion and future works}
\vspace{-3mm}
In this paper, we present an approach to use meta-learning to improve online trackers based on deep networks.  We demonstrate this by improving  two state-of-the-art trackers (CREST and MDNet). We learn to obtain a robust initial target model based on the error signals from the future frames during meta-training phase. Experimental results have shown improvements in speed, accuracy, and robustness on both trackers. The proposed technique is also general so that other trackers may benefit from it.

Other than target appearance modeling, which is the focus of this paper, there are many other important factors in object tracking algorithms. For example, when or how often to update the model~\cite{Supancic-iccv-2017}, how to manage the database~\cite{DanelljanCVPR2017}, and how to define the search space. These considerations are sometimes more important than target appearance modeling. In future work we propose including handling of these as part of learning and meta-learning.


\bibliographystyle{splncs}
\bibliography{egbib}

\newpage
\clearpage
\appendix

\section*{Appendix}
\section{More visualizations of response maps of MetaCREST}
\begin{figure}[t]
\caption{More visualizations of response maps in MetaCREST: Left three columns represents a cropped image centered on the target at the initial frame, response map with meta-learned initial correlation filters $\theta_0$, response map after updating 1 iteration with meta-learned $\alpha$, respectively. The rest of six columns on the right shows response maps of CREST after updating the model up to 10 iterations.}
\label{fig:more_vis}
\begin{center}
\includegraphics[width=\linewidth]{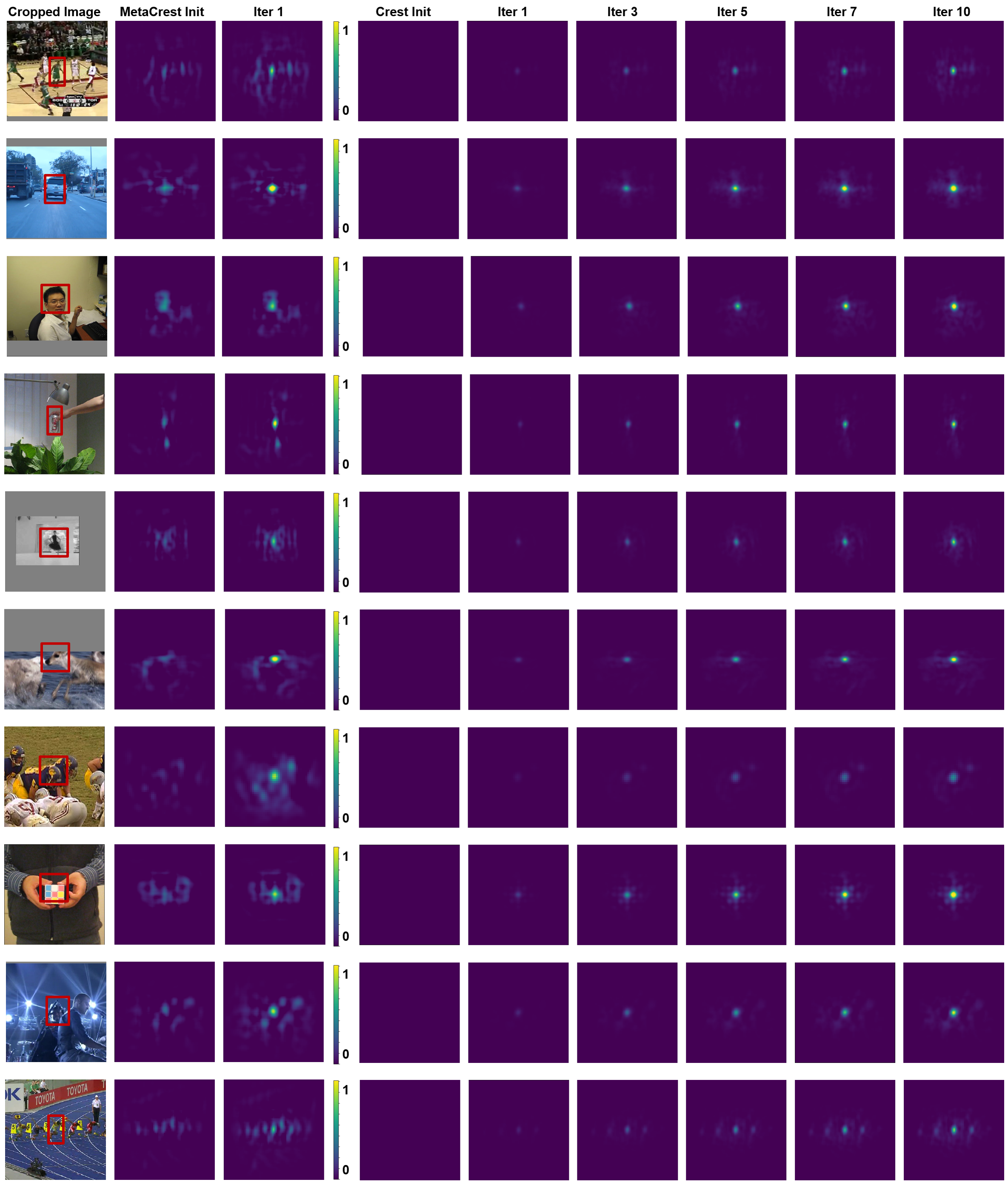}
\end{center}
\end{figure}

Figure~\ref{fig:more_vis}.

\section{Detailed results on VOT2016}
We present detailed results of MetaCREST (Table~\ref{table:vot_crest_detail_acc} and \ref{table:vot_crest_detail_rob}) and MetaSDNet (Table~\ref{table:vot_mdnet_detail_acc} and \ref{table:vot_mdnet_detail_rob}) on VOT2016 dataset. Both accuracy and robustness table are generated from VOT2016 toolkit. For original CREST tracker, we could not get the same results as reported in their paper (the performance we could get is lower). In the main text, we reported the results from their paper and we omitted detailed results of CREST since they are not available. We provided other results, CREST-Base, CREST-10, CREST-05, CREST-03, and CREST-01.

\section{Detailed results on OTB2015}
We present detailed results of MetaCREST (Table~\ref{table:otb100_crest_detail_1} and \ref{table:otb100_crest_detail_2}) and MetaSDNet (Table~\ref{table:otb100_mdnet_detail_1} and \ref{table:otb100_mdnet_detail_2}) on OTB2015 dataset. It shows the results of individual sequences in success plots.

\clearpage

\begin{table}[]
\tiny
\centering
\caption{Detailed results of MetaCREST on VOT2016 --- Accuracy.}
\label{table:vot_crest_detail_acc}
\begin{tabular}{|l|l|l|l|l|l|l|}
\hline
               & MetaCREST-01 & CREST-Base & CREST-10 & CREST-05 & CREST-03 & CREST-01 \\ \hline
bag            & 0.45         & 0.45       & 0.45     & 0.41     & 0.37     & 0.31     \\ \hline
ball1          & 0.8          & 0.8        & 0.8      & 0.81     & 0.8      & 0.8      \\ \hline
ball2          & 0.44         & 0.31       & 0.38     & 0.3      & 0.47     & 0.45     \\ \hline
basketball     & 0.6          & 0.55       & 0.55     & 0.6      & 0.62     & 0.61     \\ \hline
birds1         & 0.52         & 0.53       & 0.52     & 0.5      & 0.53     & 0.53     \\ \hline
birds2         & 0.29         & 0.28       & 0.32     & 0.28     & 0.28     & 0.28     \\ \hline
blanket        & 0.58         & 0.61       & 0.61     & 0.64     & 0.62     & 0.61     \\
\hline
\end{tabular}
\end{table}

\begin{table}[]
\tiny
\centering
\caption{Detailed results of MetaCREST on VOT2016 --- Robustness (The number of failures).}
\label{table:vot_crest_detail_rob}
\begin{tabular}{|l|l|l|l|l|l|l|}
\hline
               & MetaCREST-01 & CREST-Base & CREST-10 & CREST-05 & CREST-03 & CREST-01 \\ \hline
bag            & 0            & 0          & 0        & 0        & 0        & 0        \\ \hline
ball1          & 0            & 0          & 0        & 0        & 0        & 0        \\ \hline
ball2          & 1            & 0          & 0        & 0        & 1        & 1        \\ \hline
basketball     & 1            & 2          & 1.9      & 1        & 1        & 1        \\ \hline
birds1         & 1            & 1          & 2        & 2        & 1        & 1        \\ \hline
birds2         & 0            & 0          & 1        & 0        & 0        & 0        \\ \hline
blanket        & 0            & 0          & 0        & 0        & 0        & 0        \\
\hline
\end{tabular}
\end{table}

\begin{table}[]
\tiny
\centering
\caption{Detailed results of MetaSDNet on VOT2016 --- Accuracy.}
\label{table:vot_mdnet_detail_acc}
\begin{tabular}{|l|l|l|l|l|l|l|l|}
\hline
               & MetaSDNet-01 & pyMDNet-30 & pyMDNet-15 & pyMDNet-10 & pyMDNet-05 & pyMDNet-03 & pyMDNet-01 \\ \hline
bag            & 0.49         & 0.52       & 0.51       & 0.49       & 0.47       & 0.43       & 0.38       \\ \hline
ball1          & 0.77         & 0.79       & 0.78       & 0.79       & 0.78       & 0.77       & 0.75       \\ \hline
ball2          & 0.33         & 0.46       & 0.47       & 0.38       & 0.46       & 0.45       & 0.5        \\ \hline
basketball     & 0.62         & 0.62       & 0.63       & 0.61       & 0.63       & 0.57       & 0.53       \\ \hline
birds1         & 0.51         & 0.48       & 0.5        & 0.48       & 0.49       & 0.47       & 0.45       \\ \hline
birds2         & 0.35         & 0.33       & 0.34       & 0.35       & 0.34       & 0.31       & 0.33       \\ \hline
blanket        & 0.6          & 0.58       & 0.56       & 0.56       & 0.51       & 0.54       & 0.49       \\
\hline
\end{tabular}
\end{table}

\begin{table}[]
\tiny
\centering
\caption{Detailed results of MetaSDNet on VOT2016 --- Robustness (The number of failures).}
\label{table:vot_mdnet_detail_rob}
\begin{tabular}{|l|l|l|l|l|l|l|l|}
\hline
               & MetaSDNet-01 & pyMDNet-30 & pyMDNet-15 & pyMDNet-10 & pyMDNet-05 & pyMDNet-03 & pyMDNet-01 \\ \hline
bag            & 0            & 0.07       & 0          & 0          & 0.07       & 0.13       & 0.2        \\ \hline
ball1          & 0.4          & 0.07       & 0.13       & 0          & 0.2        & 0.33       & 1.6        \\ \hline
ball2          & 0.67         & 2.27       & 2.33       & 2.67       & 2.67       & 3.07       & 3.33       \\ \hline
basketball     & 1.27         & 1.07       & 1.27       & 1.33       & 1.8        & 3          & 5.73       \\ \hline
birds1         & 0.47         & 2.2        & 1.73       & 1.47       & 1.33       & 2.33       & 3.13       \\ \hline
birds2         & 0.4          & 0.27       & 0.4        & 0.13       & 0.13       & 0.47       & 0.6        \\ \hline
blanket        & 0            & 0          & 0          & 0          & 0.13       & 0.33       & 1          \\
\hline
\end{tabular}
\end{table}

\begin{table}[t]
\tiny
\centering
\caption{Detailed results of MetaCREST on OTB2015 (BasketBall --- Girl).}
\label{table:otb100_crest_detail_1}
\begin{tabular}{|l|l|l|l|l|l|l|l|}
\hline
             & MetaCREST-01 & CREST    & CREST-Base & CREST-20 & CREST-10 & CREST-05 & CREST-01 \\ \hline
BasketBall   & 0.658588     & 0.7194   & 0.7105     & 0.7286   & 0.7274   & 0.7302   & 0.7401   \\ \hline
Biker        & 0.262911     & 0.3414   & 0.3417     & 0.3451   & 0.3484   & 0.2495   & 0.2482   \\ \hline
Bird1        & 0.053922     & 0.3103   & 0.3306     & 0.3166   & 0.419    & 0.3757   & 0.3825   \\ \hline
Bird2        & 0.822511     & 0.7638   & 0.7605     & 0.7595   & 0.7874   & 0.7937   & 0.8278   \\ \hline
BlurBody     & 0.685201     & 0.6818   & 0.6829     & 0.6932   & 0.7039   & 0.6806   & 0.6913   \\ \hline
BlurCar1     & 0.812668     & 0.8451   & 0.8452     & 0.8451   & 0.8479   & 0.8478   & 0.8498   \\ \hline
BlurCar2     & 0.803337     & 0.8612   & 0.8623     & 0.8612   & 0.8584   & 0.8634   & 0.8593   \\
\hline
\end{tabular}
\end{table}

\begin{table}[t]
\tiny
\centering
\caption{Detailed results of MetaCREST on OTB2015 (Girl2 --- Woman).}
\label{table:otb100_crest_detail_2}
\begin{tabular}{|l|l|l|l|l|l|l|l|}
\hline
             & MetaCREST-01 & CREST    & CREST-Base & CREST-20 & CREST-10 & CREST-05 & CREST-01 \\ \hline
Girl2        & 0.594508     & 0.5784   & 0.5672     & 0.5937   & 0.6041   & 0.6046   & 0.6139   \\ \hline
Gym          & 0.512262     & 0.4946   & 0.5087     & 0.4976   & 0.5196   & 0.4995   & 0.4956   \\ \hline
Human2       & 0.63796      & 0.6762   & 0.676      & 0.6797   & 0.6932   & 0.691    & 0.6977   \\ \hline
Human3       & 0.667901     & 0.048    & 0.0482     & 0.048    & 0.0482   & 0.0492   & 0.0473   \\ \hline
Human4-2     & 0.54537      & 0.3327   & 0.3291     & 0.313    & 0.3308   & 0.3355   & 0.3542   \\ \hline
Human5       & 0.504575     & 0.5228   & 0.5215     & 0.5637   & 0.5806   & 0.3885   & 0.4031   \\ \hline
Human6       & 0.338745     & 0.2988   & 0.2934     & 0.2893   & 0.2938   & 0.2534   & 0.2137   \\
\hline
\end{tabular}
\end{table}

\begin{table}[t]
\tiny
\centering
\caption{Detailed results of MetaSDNet on OTB2015 (BasketBall --- Girl).}
\label{table:otb100_mdnet_detail_1}
\begin{tabular}{|l|l|l|l|l|l|l|l|l|}
\hline
             & MetaSDNet-01 & pyMDNet-30 & pyMDNet-15 & pyMDNet-10 & pyMDNet-05 & pyMDNet-03 & pyMDNet-01 \\
\hline
BasketBall   & 0.6004                    & 0.6386        & 0.6056        & 0.6273        & 0.7412        & 0.6121        & 0.0177        \\ \hline
Biker        & 0.3793                    & 0.339         & 0.3722        & 0.3575        & 0.3001        & 0.2884        & 0.0956        \\ \hline
Bird1        & 0.0308                    & 0.3054        & 0.2381        & 0.3228        & 0.3136        & 0.3177        & 0.0524        \\ \hline
Bird2        & 0.7196                    & 0.798         & 0.7884        & 0.7388        & 0.7253        & 0.6883        & 0.0789        \\ \hline
BlurBody     & 0.7241                    & 0.6705        & 0.7849        & 0.6806        & 0.7188        & 0.6501        & 0.696         \\ \hline
BlurCar1     & 0.8176                    & 0.7969        & 0.8145        & 0.8088        & 0.7959        & 0.0022        & 0.1475        \\ \hline
BlurCar2     & 0.8407                    & 0.8412        & 0.8528        & 0.8423        & 0.7875        & 0.6815        & 0.6893        \\
\hline
\end{tabular}
\end{table}

\begin{table}[t]
\tiny
\centering
\caption{Detailed results of MetaSDNet on OTB2015 (Girl2 --- Woman).}
\label{table:otb100_mdnet_detail_2}
\begin{tabular}{|l|l|l|l|l|l|l|l|l|}
\hline
             & MetaSDNet-01 & pyMDNet-30 & pyMDNet-15 & pyMDNet-10 & pyMDNet-05 & pyMDNet-03 & pyMDNet-01 \\
\hline
Girl2        & 0.7303          & 0.7339        & 0.7238        & 0.7257        & 0.6038        & 0.5872        & 0.6081        \\\hline
Gym          & 0.5068          & 0.5141        & 0.4567        & 0.5176        & 0.4625        & 0.4484        & 0.0254        \\\hline
Human2       & 0.7011          & 0.6262        & 0.7356        & 0.6797        & 0.6608        & 0.7351        & 0.402         \\\hline
Human3       & 0.5686          & 0.4716        & 0.0142        & 0.4545        & 0.5624        & 0.5327        & 0.0008        \\\hline
Human4-2     & 0.5927          & 0.6185        & 0.6038        & 0.567         & 0.5992        & 0.5804        & 0.0834        \\\hline
Human5       & 0.7134          & 0.624         & 0.6092        & 0.6533        & 0.7008        & 0.6225        & 0.5703        \\\hline
Human6       & 0.7504          & 0.694         & 0.7627        & 0.5205        & 0.7575        & 0.0121        & 0.003         \\
\hline
\end{tabular}
\end{table}

\end{document}